\documentclass[sigconf,authorversion]{acmart}
\usepackage{graphicx}
\usepackage{amsmath}
\usepackage{booktabs}
\usepackage{rotating}
\usepackage{framed}
\usepackage{multirow}
\usepackage{bm}
\usepackage{bbm}
\usepackage{xspace}
\usepackage{enumitem}
\usepackage{lipsum}
\usepackage{array}
\usepackage[ruled,vlined]{algorithm2e}
\usepackage{algpseudocode}

\usepackage{amsmath}
\DeclareMathOperator*{\kargmax}{k-argmax}
\DeclareMathOperator*{\argmax}{argmax}

\def \ie {\emph{i.e.}}

\def \etal {\emph{et al.}}

\newcommand{\tinytit}[1]{\noindent\textbf{#1.}}

\AtBeginDocument{%
  \providecommand\BibTeX{{%
    \normalfont B\kern-0.5em{\scshape i\kern-0.25em b}\kern-0.8em\TeX}}}

\copyrightyear{2021}
\acmYear{2021}
\setcopyright{acmcopyright}\acmConference[ICMR '21]{Proceedings of the 2021 International Conference on Multimedia Retrieval}{August 21--24, 2021}{Taipei, Taiwan}
\acmBooktitle{Proceedings of the 2021 International Conference on Multimedia Retrieval (ICMR '21), August 21--24, 2021, Taipei, Taiwan}
\acmPrice{15.00}
\acmDOI{10.1145/3460426.3463587}
\acmISBN{978-1-4503-8463-6/21/08}

\begin{document}
\fancyhead{}
\title{\textit{Learning to Select}:\\A Fully Attentive Approach for Novel Object Captioning}


\author{Marco Cagrandi, Marcella Cornia, Matteo Stefanini, Lorenzo Baraldi, Rita Cucchiara}
\email{{name.surname}@unimore.it}
\affiliation{%
  \institution{University of Modena and Reggio Emilia}
  \city{Modena}
  \country{Italy}
}
%
%
%
%
%
%

\renewcommand{\shorttitle}{\textit{Learning to Select}: A Fully Attentive Approach for Novel Object Captioning}
\renewcommand{\shortauthors}{Cagrandi, et al.}

\begin{abstract}
Image captioning models have lately shown impressive results when applied to standard datasets. Switching to real-life scenarios, however, constitutes a challenge due to the larger variety of visual concepts which are not covered in existing training sets. For this reason, novel object captioning (NOC) has recently emerged as a paradigm to test captioning models on objects which are unseen during the training phase. In this paper, we present a novel approach for NOC that learns to select the most relevant objects of an image, regardless of their adherence to the training set, and to constrain the generative process of a language model accordingly. Our architecture is fully-attentive and end-to-end trainable, also when incorporating constraints. We perform experiments on the \textit{held-out} COCO dataset, where we demonstrate improvements over the state of the art, both in terms of adaptability to novel objects and caption quality.
\end{abstract}

\ccsdesc[500]{Computing methodologies~Computer vision}
\ccsdesc[500]{Computing methodologies~Natural language generation}

\keywords{novel object captioning; region selector; constrained beam search.}


\maketitle

\section{Introduction}
Describing images has recently emerged as an important task at the intersection of computer vision, natural language processing, and multimedia, thanks to the key role it can have to empower both retrieval and multimedia systems~\cite{karpathy2015deep,anderson2018bottom,cornia2017visual,cornia2018paying,cornia2020explaining,bigazzi2020explore,stefanini2019artpedia}. Recent advances in image captioning, indeed, have demonstrated that fully-attentive architectures can provide high-quality image descriptions when tested on the same data distribution they are trained~\cite{herdade2019image,cornia2020m2,pan2020x,li2020oscar}. As the existing datasets for image captioning~\cite{young2014image,lin2014microsoft} are limited in terms of the number of visual concepts they contain, though, the application of such systems in real-life scenarios is still challenging.  
For this reason, the task of Novel Object Captioning (NOC) has recently gained a lot of attention due to its affinity towards real-world applications~\cite{hendricks16cvpr,agrawal2019nocaps,hu2020vivo}. This setting, indeed, requires a model to describe images containing objects unseen in the training image-text data, also referred to as out-of-domain visual concepts. 

Since the language model behind a NOC algorithm can not be trained to predict out-domain words, proper incorporation of such novel words during the generation phase is one of the most relevant issues in this task. Early NOC approaches~\cite{hendricks16cvpr,venugopalan17cvpr} tried to transfer knowledge from out-domain images by conditioning the model at training time on external unpaired visual and textual data. Further works~\cite{yao2017incorporating,li2019lstmp} proposed to integrate coping mechanisms in the language model to select words corresponding to the predictions of a tagger. However, these frameworks do not include a proper and explainable method to identify which objects on the scene are more relevant to be described, and consequently, lack on leveraging all the available information provided by visual inputs.
On a different line, Anderson~\etal~\cite{cbs2017emnlp} devised a Constrained Beam Search algorithm to force the inclusion of selected tag words in the output caption, following the predictions of a tagger.

Inspired by this last line of research, we combine the ability to constrain the predictions from a language model with the usage of object regions and of fully-attentive architectures, which is dominant in traditional image captioning. Precisely, we devise a model with a specific ability to select objects in the scene to be described, with a class-independent module that can work on both in-domain and out-of-domain objects. Further, we combine this with a variant of the Beam Search algorithm which can include constraints produced by the region selector, while assuring end-to-end differentiability.
We provide extensive experiments to validate the proposed approach: when tested on the \textit{held-out} portion of the COCO dataset, our model provides state-of-the-art results in terms of caption quality and adaptability to describe objects unseen in the training set. Given its simplicity and effectiveness, our approach can also be thought of as a powerful new baseline for NOC, which can foster future works in the same area.

\begin{figure*}[t]
    \centering
    \includegraphics[width=0.96\linewidth]{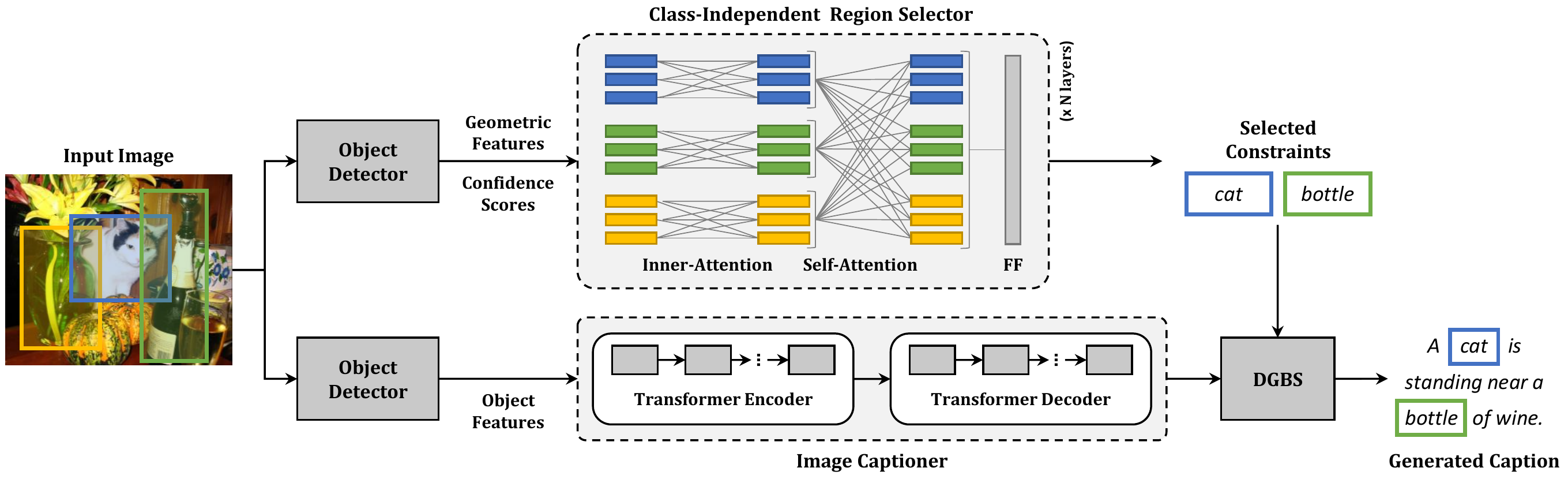}
    \vspace{-.2cm}
    \caption{Summary of our approach.}
    \label{fig:approach}
    \vspace{-.2cm}
\end{figure*}

\section{Proposed Method} \label{sec:method}
Our NOC approach can be conceptually divided into two modules: an \textit{image captioner} and a \textit{region selector}.
While the image captioning model is conditioned on the input image and is in charge of modeling a sequence of output words, the region selector is in charge of choosing the most relevant objects which need to be described, regardless of their adherence to the training set. The objects picked by the selector are used as constraints during the generation process, so that the output caption is forced to contain their labels as predicted by an object detector.
All the components of our architecture are based on fully-attentive structure, and end-to-end training is allowed also when adding constraints to the language model. Fig.~\ref{fig:approach} shows an outline of the approach.

\subsection{Class-Independent Region Selector}
The role of the region selector is to identify objects which must be described in the output sentence. Since the object selector will need to work on classes that are unseen in the training set, we adopt a class-independent strategy in which no information about the object class is employed in the feature extraction process. Instead, we model intra-class relationships between objects of the same class, to handle the case in which multiple objects of the same class are present on the scene.

Given a set of regions $\bm{X} = \{x_i\}_i$ extracted from the input image, along with their classes $\{c_i\}_i$, we extract central coordinates, width, height and, additionally, we compute the object area. We also consider as an extra feature the confidence score $s_{i}$ of the object, to obtain a class-independent feature vector:
\begin{equation}
x_{i}=\left[
\left(\frac{x_{c}}{W}\right), 
\left(\frac{y_{c}}{H}\right), 
\left(\frac{w_{i}}{W}\right), 
\left(\frac{h_{i}}{H}\right), 
\left(\frac{w_{i} \cdot h_{i}}{W \cdot H}\right),
s_{i}
\right]
\label{eq:features_geom}
\end{equation}
where $x_{c}$ and $y_{c}$ are the coordinates of the center of the region, $w_{i}$ and $h_{i}$ its width and height, and $W$ and $H$ the image dimensions.

The set of feature vectors obtained for an image is then fed to a sequence of Transformer-like~\cite{vaswani2017attention} layers, each of them composed by an \textit{inner-attention} operator and a \textit{self-attention} operator. The inner-attention operator is devised to connect together regions belonging to the same class, while the self-attention operator provides complete connectivity between elements in $\bm{X}$. The combination of these two operators allows the region selector to independently focus on specific clusters of objects, in order to exchange semantically related information and learn intra-class dependencies, and then, to model long-range and diverse dependencies.

\begin{table*}[t]
\caption{Evaluation on the \textit{held-out} COCO test set, when using different constraint selection approaches.}
\label{tab:results}
\vspace{-0.2cm}
\small
\centering
\setlength{\tabcolsep}{.35em}
\resizebox{\linewidth}{!}{
\begin{tabular}{lc cccccccc c cccccccc c cccccccc}
\toprule
& & \multicolumn{8}{c}{Cross-Entropy Loss} & & \multicolumn{8}{c}{CIDEr Optimization} & & \multicolumn{8}{c}{CIDEr Optimization with DGBS} \\
\cmidrule{3-10}
\cmidrule{12-19}
\cmidrule{21-28}
& & \multicolumn{3}{c}{In-Domain} & & \multicolumn{4}{c}{Out-Domain} & & \multicolumn{3}{c}{In-Domain} & & \multicolumn{4}{c}{Out-Domain} & & \multicolumn{3}{c}{In-Domain} & & \multicolumn{4}{c}{Out-Domain} \\
\cmidrule{3-5} \cmidrule{7-10} \cmidrule{12-14} \cmidrule{16-19} \cmidrule{21-23} \cmidrule{25-28}
& & M & C & S & & M & C & S & F1 & & M & C & S & & M & C & S & F1 & & M & C & S & & M & C & S & F1 \\
\midrule
No Constraints & & 27.2 & 108.9 & 20.2 & & 22.4 & 68.5 & 14.7 & 0.0 & & 28.4 & 122.3 & 22.3 & & 23.5 & 76.8 & 16.3 & 0.0 & & 28.1 & 120.9 & 21.9 & & 23.4 & 76.5 & 16.1 & 0.0  \\
Top-1 & & 26.2 & 97.4 & 19.2 & & 24.1 & 75.9 & 17.6 & 60.1 & & 27.6 & 110.5 & 21.3 & & 25.4 & 84.6 & 18.8 & 60.2 & & 27.9 & 115.9 & 21.0 & & 25.3 & 84.7 & 18.7 & 60.2  \\
Top-2 & & 24.4 & 81.9 & 16.4 & & 23.8 & 68.7 & 16.1 & 68.1 & & 26.2 & 95.4 & 18.4 & & 25.1 & 77.6 & 17.3 & 68.1 & & 27.1 & 102.9 & 18.4 & & 25.6 & 80.0 & 17.2 & 68.1  \\
Top-3 & & 22.7 & 69.9 & 14.4 & & 22.4 & 56.9 & 14.5 & 66.0 & & 25.1 & 83.3 & 16.5 & & 24.4 & 67.1 & 15.4 & 66.0 & & 26.6 & 92.3 & 17.0 & & 25.2 & 70.8 & 15.6 & 66.0  \\
\midrule
Region Selector (\textit{w/o Inner}) & & 25.2 & 70.6 & 17.7 & & 24.1 & 70.6 & 16.8 & 70.2 & & 26.8 & 101.5 & 19.6 & & 25.6 & 80.5 & 18.0 & 70.2 & & 27.4 & 108.0 & 19.7 & & 25.8 & 82.2 & 18.2 & 70.2 \\
\textbf{Region Selector} & & 26.2 & 97.0 & 19.2 & & \textbf{24.9} & \textbf{78.2} & \textbf{18.3} & \textbf{75.0} & & 27.6 & 109.2 & 21.1 & & \textbf{26.1} & \textbf{87.7} & \textbf{19.4} & \textbf{75.0} & & 27.9 & 115.3 & 21.0 & & \textbf{26.3} & \textbf{88.5} & \textbf{19.4} & \textbf{75.1} \\
\midrule
\midrule
\textit{Oracle Constraints} & & \textit{27.3} & \textit{107.0} & \textit{20.6} & & \textit{25.6} & \textit{84.0} & \textit{19.0} & \textit{76.0} & & \textit{28.5} & \textit{118.9} & \textit{22.5} & & \textit{26.6} & \textit{91.7} & \textit{20.2} & \textit{76.0} & & \textit{28.6} & \textit{122.9} & \textit{22.3} & & \textit{26.6} & \textit{92.3} & \textit{20.2} & \textit{76.0}  \\
\bottomrule
\end{tabular}
}
\vspace{-.2cm}
\end{table*}

Given a partition of $\bm{X}$ computed according to the class each region belongs to, \ie~$\{ \bm{r}_c \subseteq \bm{X} \mid \forall x_i, x_j \in \bm{r}_c, c_i = c_j \}_c$, the result of the inner-attention operator applied over an element of the partition is a new set of elements $\mathcal{I}(\bm{r}_c)$, with the same cardinality as $\bm{r}_c$, in which each element is replaced with a weighted sum of values computed from regions of the same class. Formally, it can be defined as:
\begin{equation}
    \mathcal{I}(\bm{r}_c) = \mathsf{Attention}(W_q\bm{r}_c, W_k\bm{r}_c, W_v\bm{r}_c),
\end{equation}
where $\bm{r}_c$ is the set of all elements of $\bm{X}$ belonging to class $c$, $W_*$ are learnable projection matrices, and $\mathsf{Attention}$ indicates the standard dot-product attention~\cite{vaswani2017attention}.

The inner attention layer is applied independently over each element of the above-defined partition so that the overall encoding of $\bm{X}$ is a new sequence of elements defined as follows:
\begin{equation}
    \mathcal{I}(\bm{X}) = \left( \mathcal{I}(\bm{r}_1), \mathcal{I}(\bm{r}_2), ..., \mathcal{I}(\bm{r}_C) \right),
\end{equation}
where $C$ indicates the number of classes. 
After each inner-attention layer, a self-attention layer is employed to connect elements of different classes together. Formally, it is defined as:
\begin{equation}
    \mathcal{S}(\bm{X}) = \mathsf{Attention}(W_q\bm{X}, W_k\bm{X}, W_v\bm{X}),
\end{equation}
where $W_*$ are, again, learnable projection matrices. 

After a sequence of inner- and self-attention layers, in which each pair of operators is followed by a position-wise feed-forward network~\cite{vaswani2017attention}, the region selector outputs a selection score $Y_{i}$ for each object proposal. To do so, we apply an affine transformation and a non-linear activation to the output of the last layer:
\begin{equation}
Y_{i} = \sigma\left(\mathsf{RegionSelector}(X_{i})W_{o}\right),
\end{equation}
where $W_{o} \in \mathbf{R}^{d \times 1}$ are learnable weights and $\sigma$ is a sigmoid.

\tinytit{Training} The region selector is trained using a binary cross-entropy loss. To build ground-truth data, for each image we collect the object classes identified by the object detector and construct a binary ground-truth vector indicating whether a class name is contained in at least one of the ground-truth captions associated with the image.
We also consider as positives synonyms and plural forms of the object class names.
At inference time, we extract the selected objects for each image adopting $0.5$ as threshold.

\subsection{Image Captioner}
After object selection, our image captioning model is responsible for generating a caption using the chosen class names as constraints. Inspired by recent works which employ fully-attentive models in image captioning~\cite{herdade2019image,cornia2020m2,luo2021dual}, we create a captioning model with an encoder-decoder structure, where the encoder refines image region features and the decoder generates captions auto-regressively. 

\tinytit{Encoder}
Recent captioning literature has shown that object regions are the leading solution to encode visual inputs~\cite{anderson2018bottom,yao2018exploring,yang2019auto}, followed by self-attentive layers to model region relationships~\cite{herdade2019image,huang2019attention,cornia2020m2,pan2020x,stefanini2020novel,luo2021dual}. However, as self-attention can only encode pairwise similarities, it exhibits a significant limitation on encoding knowledge learned from data. 
To overcome this restraint, we enrich our encoder with memory slots~\cite{cornia2020smart,cornia2020m2}. Specifically, we extend the set of keys and values of self-attention layers with additional learnable vectors, which are independent of the input sequence and can encode a priori information retrieved through attention.

\tinytit{Decoder}
The decoder is the actual language model, conditioned on both previously generated words and image region encodings.
As in the standard Transformer~\cite{vaswani2017attention}, our language model is composed of a stack of decoder layers, each performing a masked self-attention and a cross-attention followed by a position-wise feed-forward network.
Specifically, for each cross-attention, keys and values are inferred from the encoder output, while for the masked self-attention, queries, keys, and values are exclusively extracted from the input sequence of the decoder. This self-attention is right-masked so that each query can only attend to keys obtained from previous words.

\subsection{Including Lexical Constraints}
To include the lexical constraints produced by the region selector when decoding from the language model, we devise a variant of the Beam Search algorithm~\cite{och2004alignment,hokamp2017lexically} which supports the adoption of single-word constraints. Given a number of word constraints $\bm{W} = \{w_0, w_1, ..., w_n\}$ and a maximum decoding length $T$, we frame the decoding process in a matrix $\bm{G}$ with $n$ rows and $T$ columns, where the horizontal axis covers the time steps in the output sequence, and the vertical axis indicates the constraints coverage. Each cell of the matrix can contain a beam of partially decoded sequences.

At iteration $t$, each row $i$ of $\bm{G}[:, t]$ can be filled in two ways: either by continuing the beam contained in $\bm{G}[i,t-1]$ by sampling from the probability distribution of the language model, or by forcing the inclusion of a constraint from $\bm{W}$. In the former case, the resulting updated beam of sequences is stored in $\bm{G}[i,t]$, while in the latter case it is stored in $\bm{G}[i+1,t]$. At the end of the generation process, the last row of $\bm{G}$ will contain sequences that satisfy all constraints.

Algorithm~\ref{algo:gbs} reports the pseudo-code of our constrained beam search procedure. There, $k$ indicates the number of elements in each bin, $\text{model.step}$ indicates sampling from the language model probability distribution to continue the generation of a partially-decoded sequence, while $\text{model.add\_constr}$ indicates a function which continues a beam by adding all the available constraints, excluding those which have already been generated for a sequence.
Because all the operations required to include constraints are differentiable, we call our constraint inclusion approach \textit{Differentiable Grid Beam Search} (DGBS), and employ it to fine-tune the image captioner also when using a CIDEr-D optimization strategy.

\begingroup
\setlength{\textfloatsep}{5pt}
\algrenewcommand{\algorithmiccomment}[1]{\hspace{1em}$\triangleright$ #1}
\begin{algorithm}[tb]
\footnotesize
\SetAlgoLined
 $\bm{G} \leftarrow \text{initGrid}(n,T,k)$
 
 \For{$t=1; t<T; t++$}{
  \For{$c=\max(0, n+t-T); c<\min(t, n); c++$}{
  $g,s = \emptyset$
  \ForAll{hyp in $\bm{G}[c,t-1]$}{
  $g\leftarrow g \cup \text{model.step}(hyp)$
  }
  \If{$c > 0$}{
      \ForAll{hyp in $\bm{G}[c-1,t-1]$}{
      $s\leftarrow s \cup \text{model.add\_constr}(hyp, \{w_0, ..., w_n\})$ 
      }
  }
  $\bm{G}[c,t] \leftarrow \underset{h \in g \cup s}{\kargmax}\left( \text{model.score}(h) \right)$
  }
 }
 $topHyp \leftarrow \text{hasEOS}(\bm{G}[n,:])$\Comment{\small{Remove sequences w/o EOS}}
 
 \Return $\underset{h \in topHyp}{\argmax}\left( \text{model.score}(h) \right)$
 \caption{Grid Beam Search}
 \label{algo:gbs}
\end{algorithm}
\endgroup

\begin{table*}[t]
\caption{Comparison with the state of the art on the \textit{held-out} COCO test set.}
\label{tab:sota_comparison}
\vspace{-0.2cm}
\resizebox{0.85\linewidth}{!}{
\footnotesize
\centering
\setlength{\tabcolsep}{.35em}
\begin{tabular}{lc ccccccccc c ccccc}
\toprule
& & \multicolumn{9}{c}{F1 Scores} & & \multicolumn{5}{c}{Captioning Metrics} \\
\cmidrule{3-11}
\cmidrule{13-17}
 & & F1$_\text{bottle}$ & F1$_\text{bus}$ & F1$_\text{couch}$ & F1$_\text{microwave}$ & F1$_\text{pizza}$ & F1$_\text{racket}$ & F1$_\text{suitcase}$ & F1$_\text{zebra}$ & F1$_\text{average}$ & & B-4 & M & R & C & S \\
\midrule
DCC~\cite{hendricks16cvpr} & & 4.6 & 29.8 & 45.9 & 28.1 & 64.6 & 52.2 & 13.2 & 79.9 & 39.8 & & - & 21.0 & - & 59.1 & 13.4 \\
NOC~\cite{venugopalan17cvpr} & & 17.8 & 68.8 & 25.6 & 24.7 & 69.3 & 55.3 & 39.9 & 89.0 & 48.8 & & - & 21.3 & - & - & - \\
NBT~\cite{lu2018neural} & & 14.0 & 74.8 & 42.8 & 63.7 & 74.4 & 19.0 & 44.5 & 92.0 & 53.2 & & - & 23.9 & - & 84.0 & 16.6 \\
CBS~\cite{cbs2017emnlp} & & 16.3 & 67.8 & 48.2 & 29.7 & 77.2 & 57.1 & 49.9 & 85.7 & 54.0 & & - & 23.6 & - & 77.6 &  15.9 \\
LSTM-C~\cite{yao2017incorporating} & & 29.7 & 74.4 & 38.8 & 27.8 & 68.2 & 70.3 & 44.8 & 91.4 & 55.7 & & - & 23.0 & - & - & - \\
DNOC~\cite{wu2018dnoc} & & 33.0 & 76.9 & 54.0 & 46.6 & 75.8 & 33.0 & 59.5 & 84.6 & 57.9 & & - & 21.6 & -& - & - \\
LSTM-P~\cite{li2019lstmp} & & 28.7 & 75.5 & 47.1 & 51.5 & 81.9 & 47.1 & 62.6 & 93.0 & 60.9 & & - & 23.4 & - & 88.3 & 16.6  \\
NBT + CBS~\cite{lu2018neural} & & 38.3 & 80.0 & 54.0 & \textbf{70.3} & 81.1 & 74.8 & 67.8 & \textbf{96.6} & 70.3 & & - & 24.1 & - & 86.0 & 17.4 \\
\midrule
Top-2 & & 29.6 & 77.4 & 44.7 & 62.6 & 83.3 & \textbf{81.2} & 70.7 & 95.1 & 68.1 & & 28.1 & 25.6 & 52.7 & 80.0 & 17.2 \\
Region Selector (\textit{w/o Inner}) & & 42.3 & 78.3 & 54.4 & 59.4 & 85.3 & 79.1 & 67.2 & 95.6 & 70.2 & & 28.4 & 25.8 & 52.8 & 82.2 & 18.2 \\
\textbf{Region Selector} & & \textbf{43.9} & \textbf{83.7} & \textbf{66.8} & 64.7 & \textbf{88.0} & 81.0 & \textbf{76.9} & 95.4 & \textbf{75.1} & & \textbf{30.3} & \textbf{26.3} & \textbf{53.8} & \textbf{88.5} & \textbf{19.4} \\
\bottomrule
\end{tabular}
}
\vspace{-.1cm}
\end{table*}

\begin{table}[t]
\caption{Region selector performance evaluation using different loss weights for zero and one values.}
\label{tab:ablation}
\vspace{-0.2cm}
\footnotesize
\centering
\setlength{\tabcolsep}{.30em}
\resizebox{\linewidth}{!}{
\begin{tabular}{lcccc ccc c cccc}
\toprule
& & & & & \multicolumn{3}{c}{In-Domain} & & \multicolumn{4}{c}{Out-Domain} \\
\cmidrule{6-8} \cmidrule{10-13}
& & $\lambda_0$ & $\lambda_1$ & & M & C & S & & M & C & S & F1 \\
\midrule
Region Selector (\textit{w/o Inner})  & & 0.4 & 0.6 & & 27.4 & 111.9 & 20.3 & & 25.8 & 85.6 & 18.7 & 68.5 \\
\textbf{Region Selector} & & 0.4 & 0.6 & & \textbf{28.1} & \textbf{119.2} & \textbf{21.3} & & \textbf{26.0} & \textbf{89.0} & \textbf{19.4} & \textbf{70.4} \\
\midrule
Region Selector (\textit{w/o Inner}) & & 0.3 & 0.7 & & 27.2 & 108.7 & 19.9 & & 25.9 & 84.9 & 18.6 & 69.9 \\
\textbf{Region Selector} & & 0.3 & 0.7 & & \textbf{28.0} & \textbf{116.4} & \textbf{21.2} & & \textbf{26.2} & \textbf{88.7} & \textbf{19.4} & \textbf{74.2} \\
\midrule
Region Selector (\textit{w/o Inner}) & & 0.2 & 0.8 & & 27.4 & 108.0 & 19.7 & & 25.8 & 82.1 & 18.2 & 70.2 \\
\textbf{Region Selector} & & 0.2 & 0.8 & & \textbf{27.9} & \textbf{115.3} & \textbf{21.0} & & \textbf{26.3} & \textbf{88.5} & \textbf{19.4} & \textbf{75.1} \\
\midrule
Region Selector (\textit{w/o Inner}) & & 0.1 & 0.9 & & 26.9 & 97.8 & 18.2 & & 25.6 & 73.2 & 16.7 & 67.1 \\
\textbf{Region Selector} & & 0.1 & 0.9 & & \textbf{27.9} & \textbf{114.3} & \textbf{20.8} & &  \textbf{26.2} & \textbf{87.5} & \textbf{19.2} & \textbf{75.6} \\
\bottomrule
\end{tabular}
}
\vspace{-.3cm}
\end{table}

\section{Experiments} \label{sec:results}
\subsection{Evaluation Protocol}
\tinytit{Dataset} We conduct experiments on the \textit{held-out} COCO dataset~\cite{hendricks16cvpr}, which consists of a subset of the COCO dataset~\cite{lin2014microsoft} for standard image captioning, 
where the training set excludes all image-caption pairs that mention at least one of the following eight objects: \textit{bottle}, \textit{bus}, \textit{couch}, \textit{microwave}, \textit{pizza}, \textit{racket}, \textit{suitcase}, and \textit{zebra}. We follow the splits defined in~\cite{hendricks16cvpr} and take half of COCO validation set for validation and the other half for testing.

\tinytit{Metrics} To evaluate caption quality, we use standard captioning metrics (\ie~BLEU-4~\cite{papineni2002bleu}, METEOR~\cite{banerjee2005meteor}, ROUGE~\cite{lin2004rouge}, CIDEr~\cite{vedantam2015cider}, and SPICE~\cite{spice2016}), while we employ F1-scores~\cite{hendricks16cvpr} to measure the model ability to incorporate new objects in generated captions. 

\tinytit{Implementation details}
To extract geometric features and confidence scores for our region selector, we employ Faster R-CNN~\cite{ren2017faster} with ResNet-50-FPN backbone, trained on COCO~\cite{lin2014microsoft}. For both training and inference, we discard the detections of the \textit{person} and \textit{background} classes. 
During training, we use different loss weights (\ie,~$\lambda_0 = 0.2$ and $\lambda_1=0.8$) to balance the importance of zero and one ground-truth values, and we limit the number of object proposals for each image to $10$ according to their confidence scores. 
Region selector features are projected to a $128$-dimensional embedding space and passed through $N=2$ identical layers, each composed of inner-attention, self-attention, and feed-forward. 

For our image captioning model, we extract object features from Faster R-CNN~\cite{ren2017faster} with ResNet-101 finetuned on Visual Genome~\cite{krishnavisualgenome,anderson2018bottom}. Following~\cite{cornia2020m2}, we use three layers for both encoder and decoder and employ $40$ memory vectors for each encoder layer. We represent words with GloVe word embeddings~\cite{pennington2014glove}, using two fully-connected layers to convert between the GloVe dimensionality (\ie,~300) and the captioning model dimensionality (\ie,~512) before the first decoding layer and after the last decoding layer. Before the final softmax, we multiply with the transpose of the word embeddings.
We pre-train our captioning model using cross-entropy and finetune it using RL with CIDEr-D reward. During this phase, we use the classes detected by Faster R-CNN, trained on COCO, that are mentioned in the ground-truth captions as constraints for our DGBS algorithm. We limit the number of possible constraints to $5$.

All experiments are performed with a batch size equal to $50$. For training the region selector and pre-training the captioning model, we use the learning rate scheduling strategy of~\cite{vaswani2017attention} with a warmup equal to $10,000$ iterations and Adam~\cite{kingma2014adam} as optimizer. CIDEr-D optimization is done with a learning rate equal to $5\times10^{-6}$.

\subsection{Experimental Results}
Table~\ref{tab:results} shows the results of our model in terms of captioning metrics and F1-score averaged over the eight held-out classes, using different strategies to train the captioning model. We compare with a variant of our region selector without inner-attention (\ie,~\textit{w/o Inner}) and using the top-$k$ detections, with $k=1,2,3$, instead of our selection strategy. For reference, we also report the performance when using oracle constraints coming from ground-truth captions. As it can be seen, our solution achieves the best results in terms of both caption quality and F1-score, demonstrating the effectiveness of our region selector for choosing constraints for the captioning model and the importance of the inner-attention operator. Furthermore, by comparing the results with standard CIDEr optimization and those obtained using our DGBS algorithm during training, we can see improved results on both in-domain and out-domain captions, thus confirming the usefulness of our training strategy. 

In Table~\ref{tab:ablation}, we show the results when using different weights to balance the importance of zero and one ground-truth values. As it can be seen, our complete region selector achieves better performance than the variant without inner-attention, thus further demonstrating the effectiveness of the proposed attention operator. Additionally, employing $\lambda_0=0.2$ and $\lambda_1=0.8$ provides the best balance in terms of captioning metrics and F1-score.

Finally, in Table~\ref{tab:sota_comparison}, we compare our model with NOC state-of-the-art approaches. As it can be noticed, our region selector obtains the best results in terms of both F1-scores and captioning metrics, achieving a new state of the art on the \textit{held-out} COCO dataset.

\section{Conclusion}
We have presented a fully-attentive approach for NOC that learns to select and describe unseen visual concepts. Our method is based on a class-independent region selector and an image captioning model trained with a differentiable grid beam search algorithm that generates sentences with given constraints, in an end-to-end fashion. Experimental results have showed that our model achieves a new state of the art on the \textit{held-out} COCO dataset, demonstrating its effectiveness in successfully describing novel objects.




\begin{acks}
This work has been supported by ``Fondazione di Modena'', by the national project ``IDEHA: Innovation for Data Elaboration in Heritage Areas'' (PON ARS01\_00421), cofunded by the Italian Ministry of University and Research, and by the project ``Artificial Intelligence for Cultural Heritage (AI for CH)'', cofunded by the Italian Ministry of Foreign Affairs and International Cooperation.
\end{acks}


\bibliographystyle{ACM-Reference-Format}
\bibliography{bibliography}

%
%
%

\end{document}